\documentclass[runningheads]{llncs}
\usepackage[T1]{fontenc}
\usepackage{graphicx}
\usepackage{cite}
\usepackage{amsmath,amssymb,amsfonts}
\usepackage{algorithmic}
\usepackage{graphicx}
\usepackage{textcomp}
\usepackage{xcolor}
\usepackage{booktabs}
\usepackage{xcolor}
\usepackage{multirow,multicol,placeins}
\usepackage[marginal]{footmisc}
\usepackage{subfig}

\begin{document}
\title{Push the Limit of Multi-modal Emotion Recognition by Prompting LLMs with Receptive-Field-Aware Attention Weighting}
\titlerunning{Lantern}

\author{%
  Han Zhang\inst{1*} \and
  Yu Lu\inst{1*} \and
  Liyun Zhang\inst{1} \and
  Dian Ding\inst{1\#} \and
  Dinghua Zhao\inst{1} \and
  Yi-Chao Chen\inst{1} \and
  Ye Wu\inst{2} \and
  Guangtao Xue\inst{1\#}
}

\authorrunning{Han Zhang, Yu Lu et al.}

\institute{%
  Shanghai Jiao Tong University, Shanghai, China \and
  Xi'an Jiaotong-Liverpool University, Suzhou, China
}

%
%
\maketitle              
\footnote{*These authors contribute equally to this work.\\
$^\#$Corresponding author. E-mails:\email{dingdian94@sjtu.edu.cn}, \email{gt\_xue@sjtu.edu.cn}}

\newcommand{\sysname}{\emph{Lantern}}.
\begin{abstract}
Understanding the emotions in a dialogue usually requires external knowledge to accurately understand the contents. 
As the LLMs become more and more powerful, we do not want to settle on the limited ability of the pre-trained language model. 
However, the LLMs either can only process text modality or are too expensive to process the multimedia information. 
We aim to utilize both the power of LLMs and the supplementary features from the multimedia modalities. 
In this paper, we present a framework, \sysname{}, that can improve the performance of a certain vanilla model by prompting large \underline{\textbf{lan}}guage models with recep\underline{\textbf{t}}ive-fi\underline{\textbf{e}}ld-awa\underline{\textbf{r}}e atte\underline{\textbf{n}}tion weighting. 
This framework trained a multi-task vanilla model to produce probabilities of emotion classes and dimension scores. 
These predictions are fed into the LLMs as references to adjust the predicted probabilities of each emotion class with its external knowledge and contextual understanding. 
We slice the dialogue into different receptive fields, and each sample is included in exactly $t$ receptive fields. 
Finally, the predictions of LLMs are merged with a receptive-field-aware attention-driven weighting module. 
In the experiments, vanilla models CORECT and SDT are deployed in \sysname{} with GPT-4 or Llama-3.1-405B. 
The experiments in IEMOCAP with 4-way and 6-way settings demonstrated that the \sysname{} can significantly improve the performance of current vanilla models by up to 1.23\% and 1.80\%. 

\end{abstract}

\section{Introduction}
\label{sec:intro}
Understanding human emotions is always considered one of the necessary qualities for a powerful AI. 
Nevertheless, emotion recognition is a tricky task. It requires the models to understand the contents of dialogues, analyze the tones in the acoustic signals, and capture the expressions in the videos. 
Recent studies \cite{nguyen2023conversation,joshi2022cogmen} have made remarkable progress. These studies focus on how we extract information from each modality and fuse them to better understand the emotion efficiently. 
However, some dialogues need certain knowledge about modern ethics to infer the correct emotions. One example of this demand is sarcasm, which often has an emotion against the literal meanings. 
Most current studies only train models from scratch with specific emotion recognition corpora. 
Although these corpora have thousands of samples, the models are still hard to learn and understand the underlying ethical knowledge. 

\begin{figure}[h]
    \centering
    \includegraphics[ width=0.8\textwidth]{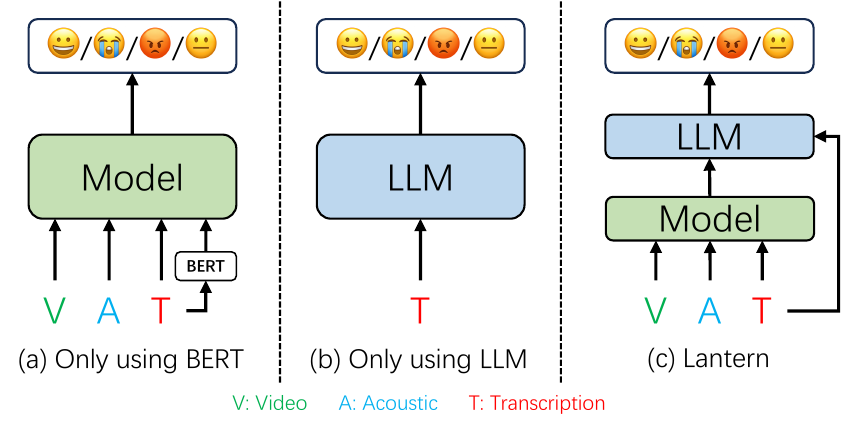}
    \caption{Integrating external knowledge: The first approach uses a pre-trained language model to generate a new modality from text with external knowledge. The second approach leverages an LLM to process dialogue transcriptions. Our framework uses a vanilla model to process multimedia modalities and generate intermediate supportive information for LLM refinement.}
    \label{fig:cmp_ppl}
\end{figure}

Integrate a pre-trained natural language model \cite{wu2021emotion, Fang2024Pre_ppnallm} to pre-process the text modality with its knowledge has been widely used to increase the model's awareness of external knowledge.
As shown in Figure \ref{fig:cmp_ppl}a, these frozen models are pre-trained on large corpora that are not specific for emotion recognition. 
They could have learned more comprehensive knowledge to understand the dialogues accurately. 
Previous studies built a vanilla model leveraging the BERT \cite{devlin2019bert} to provide external knowledge for the backbone model for understanding the context. 
However, the shortage is that given the limited processability and context window of the BERT, only very limited sentences can be sent to BERT simultaneously. 
Lacking the information of the dialogues could cause the BERT to extract misdirected features. 

Large language models have become a new trend in the study of natural language processing. 
These models with extremely numerous parameters are trained on a great deal of text corpora. 
They mastered more profound knowledge about the knowledge of our world and society and provided a larger context window compared to the BERT. 
Regarding the text inputs, hardly any vanilla models can outperform the LLMs. 
Nonetheless, most of the LLMs, like GPT-3.5, can only process text modality. 
Even if the latest GPT-4o could process multi-modal inputs, asking LLMs to process these multimedia modalities is costly. 
For example, a typical video in IEMOCAP \cite{busso2008iemocap} is 720$\times$480, which would cost 425 tokens for one frame when using GPT-4o. A 30-second video with 24 frames per second is over 300k tokens, while the context window size of GPT-4o is only 128k. 
It is still unacceptable even if we compress it to 8 frames per second with a total of 100k tokens. 
In addition, from the ablation studies from \cite{nguyen2023conversation} and \cite{joshi2022cogmen}, we find that the improvement brought by the video modality is not significant enough for us to bear the cost. 
Text is the most important modality in emotion recognition. 

In summary, while BERT is efficient, it is limited in context window and knowledge. 
Contrarily, LLMs are powerful but prohibitively expensive for processing multimedia information. 
We have been exploring a solution that can efficiently incorporate external knowledge as powerful as LLMs while still supporting multimedia modalities efficiently. 

\begin{figure*}[htb!]
    \centering
    \includegraphics[width=0.95\linewidth]{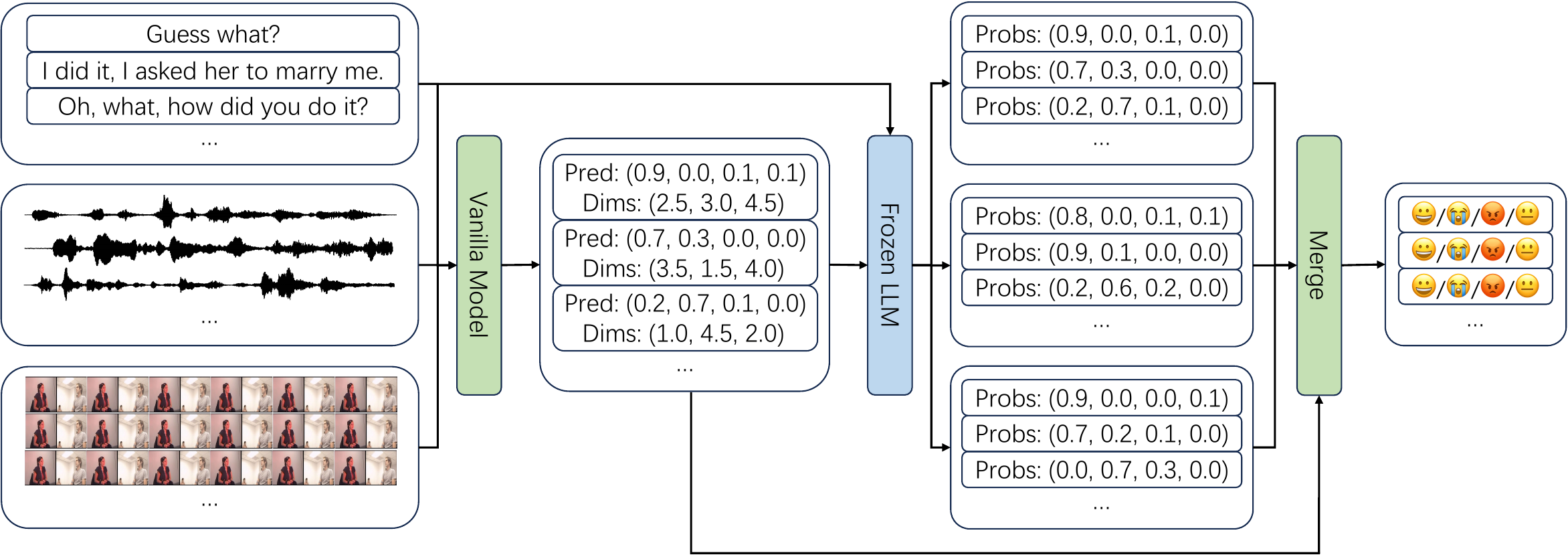}
    \caption{Our framework \sysname{} has three stages: first, a vanilla model pre-processes all modalities to produce a preliminary prediction of the probabilities and the dimension scores for each sample. Second, prompts with the preliminary predictions and the transcriptions are fed into a frozen LLM for further adjustment. Each sample will be included in $t$ prompts. Finally, a receptive-field-aware attention algorithm is implemented to assign weights for the $t+1$ predictions to form the final prediction. }
    \label{fig:pipeline}
\end{figure*}

In this paper, we present a new framework, \textbf{\sysname{}}, prompting large \underline{\textbf{lan}}guage models with recep\underline{\textbf{t}}ive-fi\underline{\textbf{e}}ld-awa\underline{\textbf{r}}e atte\underline{\textbf{n}}tion weighting. 
As shown in Figure \ref{fig:cmp_ppl}c, \sysname{} uses a vanilla model to pre-process multimedia modalities and generate initial predictions. 
These predictions are then adjusted using a \textbf{frozen LLM} with its external knowledge and contextual understanding. 

\sysname{} poses several challenges. First, we need to deliver information from multimedia modalities to LLMs efficiently.
Besides the classification metric used by most of the studies, dimension scores \cite{russell1977evidence} are another option to evaluate emotions. 
According to the results from \cite{wagner2023dawn} and \cite{atmaja2020dimensional}, acoustic signals play a more important role in predicting the dimension scores than linguistic information. 
Therefore, we deliver emotion classification metrics and dimension scores as supplementary information correlated with multimedia modalities for the LLMs' reference.
Furthermore, multitasking vanilla models are applied for both the categorical metrics and dimensional scores to enable a lightweight pipeline.

As powerful as LLMs are, their context window is still limited. 
In practice, when the length of a dialogue is out of its capability, we must split the contents. 
We take the sliding window strategy as the splitting policy. 
Each sample will be included in exactly $t$ windows to reduce the randomness of the LLM responses. 
The samples are contained in different windows to provide the LLM with different receptive fields to mine the local and overall features of the dialogue.
This feature enables us to design an attention-based algorithm to assign weights for each prediction considering the length and reliability of receptive fields. 

\sysname{} deploys a multi-task vanilla model based on CORECT \cite{nguyen2023conversation} and SDT \cite{ma2023transformer}, while using GPT-4 \cite{achiam2023gpt} and Llama-3.1-405B \cite{dubey2024llama} as the LLM agent.
A single NVIDIA RTX 3090 GPU is enough to train a vanilla multi-modal model and a lightweight merge model. 
We conducted experiments on the most commonly used dataset IEMOCAP \cite{busso2008iemocap}. The results confirm that \sysname{} significantly improves the vanilla model recognition accuracy by 1.23\% to 1.80\% in 6-way and 4-way settings when using CORECT and 0.79\% when using SDT in the 6-way setting. 

Overall, our contributions are as follows:
\begin{itemize}
    \item We present an LLM-assisted emotion recognition framework that utilizes a multitask vanilla model to extract emotion classification metrics and dimension scores from multi-modal samples, while efficiently incorporating external knowledge through the use of an LLM.
    
    \item We utilize a sliding window strategy to segment long dialogues, provide multiple receptive fields for the samples, and address the limitation of the LLM  context window.
     
    \item We employ a receptive-field-aware attention-driven weighting mechanism that assigns appropriate weights to each prediction. Compared with the prediction results of vanilla, the accuracy of sentiment recognition by fusing vanilla and LLM is significantly improved.
\end{itemize}

\section{Related Work}
\label{sec:related}
\textbf{Emotion recognition. }
The emotion recognition task requires algorithms to analyze the speakers' emotions from the transcriptions, acoustic signals, and videos. There are two ways to measure the emotion of a speaker: discrete emotions, such as happy, sad, and angry, and emotion dimensions, which describe emotions through arousal, valence, and dominance.
IEMOCAP \cite{busso2008iemocap} provided the transcriptions, acoustic signals, and videos as the input, and both the discrete emotion and dimension labels. 
\textbf{Vanilla multi-modal models. }
Different modalities can supplement each other by providing more information or confirming shared information to help models better understand the world \cite{baltruvsaitis2018multimodal, majumder2018multimodal}. Convolutional Neural Networks (CNNs) \cite{Jian2020A_ppna} has been extensively applied in emotion recognition tasks. 
The transformer structure \cite{vaswani2017attention} has significantly improved the performance of fusing multiple modalities. 
It is often used to handle the visual and language features \cite{xu2023multimodal}. 
\cite{shi2023multiemo} and \cite{delbrouck2020transformer} have applied the transformer to this task and achieved great success. 

Many researchers choose to use a Graph Neural Network (\cite{gori2005new}, \cite{busso2008iemocap}) to capture the relationship in a conversation. 
The GNN has different options for its base structure. The most commonly used structures are convolution \cite{kipf2016semi}, recurrence \cite{nicolicioiu2019recurrent}, and attention \cite{velickovic2018graph}. 
\cite{nguyen2023conversation}, \cite{joshi2022cogmen}, and \cite{hu2021dialoguecrn} deployed GNN to model the conversation to achieve significant results. 
\cite{hu2021dialoguecrn}, \cite{ghosal2020cosmic}, \cite{majumder2019dialoguernn}, and \cite{hazarika2018icon} use the Gated Recurrent Unit (GRU) as an alternative method. 

Though the studies above focused on the classification task, \cite{wagner2023dawn}, \cite{srinivasan2022representation}, and \cite{atmaja2020dimensional} explored the methods to predict the dimension scores of emotions. 
Considering the importance of acoustic signals in the task of dimension scores, they applied the wav2vec 2.0 \cite{baevski2020wav2vec} and HuBERT \cite{hsu2021hubert} to process the speech. 
\textbf{Prompt LLMs for downstream tasks. } 
The foundation of LLMs is the scaling law \cite{kaplan2020scaling}, which discovered that the models can achieve remarkable performance improvement by training with an enormously large dataset. 
For example, Llama 3 \cite{touvron2023llama} is trained on 15T tokens. With a large amount of training data, the LLMs are well-learned to understand the context of conversations with their internal knowledge. 
Therefore, it is intuitive that many works requiring external knowledge have included LLMs as part of their pipelines. 
In Visual Question Answering (VQA) tasks, \cite{shao2023prompting} used vanilla local models to generate heuristics-enhanced prompts to provide information on images for LLMs. 
\cite{zhang2023dialoguellm} uses GPT to generate video description and context knowledge for a fine-tuned Llama model for further prediction. 

\section{Method}\label{sec:method}

Our framework consists of three stages depicted in Figure \ref{fig:pipeline}. In the first stage, we use a vanilla model to process the transcriptions and multi-modal inputs to produce preliminary predictions as the inputs of the LLMs. In the second stage, we generate the prompts for the LLMs to ask them to make adjustments according to our predictions in the first stage and their knowledge. In the end, we use a receptive-field-aware attention weighting algorithm to calculate the weighted sum of each prediction to form the final results. 

\subsection{Adaption for Vanilla Models} \label{sec:vanilla}

\begin{figure}[h]
    \centering
    \includegraphics[width=0.8\linewidth]{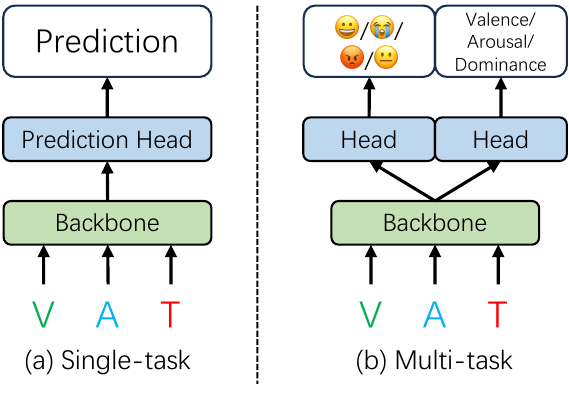}
    \caption{Methods to predict metrics: Figure \ref{fig:vanilla}a described a single-task model, where a dedicated backbone is used to extract specific to a task. Figure \ref{fig:vanilla}b demonstrated a multi-task pattern, where backbone extract features suitable for both metrics and the predictions are based on the same feature. }
    \label{fig:vanilla}
\end{figure}

The vanilla models should produce the classification predictions and dimension scores as sources of the prompts. 
However, current studies usually only focus on one of the tasks. 
These models shared the idea of using a backbone network to extract the features as a high-dimension vector as seen in the Figure \ref{fig:vanilla}a. 
The result type depends on the prediction head and the labels. 

The classification and dimension scores are both descriptions of the emotion and they are correlative variables (demonstrated in next section). It is trivial to extract the features twice with two different models. 
Therefore, as shown in Figure \ref{fig:vanilla}b, for a given vanilla model, an additional linear layer is added after the backbone to allow it to predict both tasks from the same emotion feature. 
This multi-task training pattern can reduce the burden at this step and provide a more comprehensive view during the training of the backbone network. 
The backbone of the vanilla models can be any emotion recognition model. In this work, we adopted the CORECT \cite{nguyen2023conversation} as our backbone. 

\subsection{Prompting LLMs} \label{sec:prompt}

A prompt for the LLMs has two parts, the header and the samples. 
Each sample includes the transcription, the predicted dimension scores, and the preliminary prediction of the classification probabilities. 
The header describes the task, inputs, and output format. 
Each sample is repeated $t$ times to reduce the randomness in the output of LLMs. 

The form of the task can have two options, asking the LLMs to directly produce the most likely class for each sample or the probabilities for each class. 
The procedure first option implied in the first option is asking the LLMs to produce the probabilities and output the class of highest probabilities. 
It includes one more step in the reasoning chain of the LLMs, which is unnecessary and both depth reasoning and mathematical thoughts can lead to worse performance. 
In addition, we can only use the voting algorithm to merge the $t$ predictions, which could cause coarser granularity and higher chances of tying. 
Due to these reasons, we ask the LLMs to adjust the preliminary probabilities generated by vanilla models and add the probabilities of each emotion when combining the $t$ predictions. 

To help the LLMs have a more accurate understanding of the relationship between the dimension scores and the emotion classes, we summarized the statistics of the dimension scores of each emotion (detailed in next section), which is appended in the header for the LLMs' reference. 
Inspired by \cite{wei2022chain}, we selected a series of examples from the training set and manually constructed answers for them. 
These examples are attached for the LLMs to better understand the description. 
These answers guide the LLMs to not only examine the meaning of the transcriptions, the values of class probabilities, and the dimension scores but also compare the changes between samples. 

Another step we adopted to reduce the randomness of LLMs' response is filtering the low-quality ones with an integrity check including: 1). whether all samples provided to the LLMs are predicted, 
2). whether each sample's response contains all emotions and 3). whether the sums of the adjusted probabilities are 1. 
Any response that fails to pass the integrity check will be discarded, and the LLMs will be required to respond to these samples again. 

\subsection{Split Strategy} \label{sec:split}

\begin{figure}[h]
    \centering
    \includegraphics[width=0.8\linewidth]{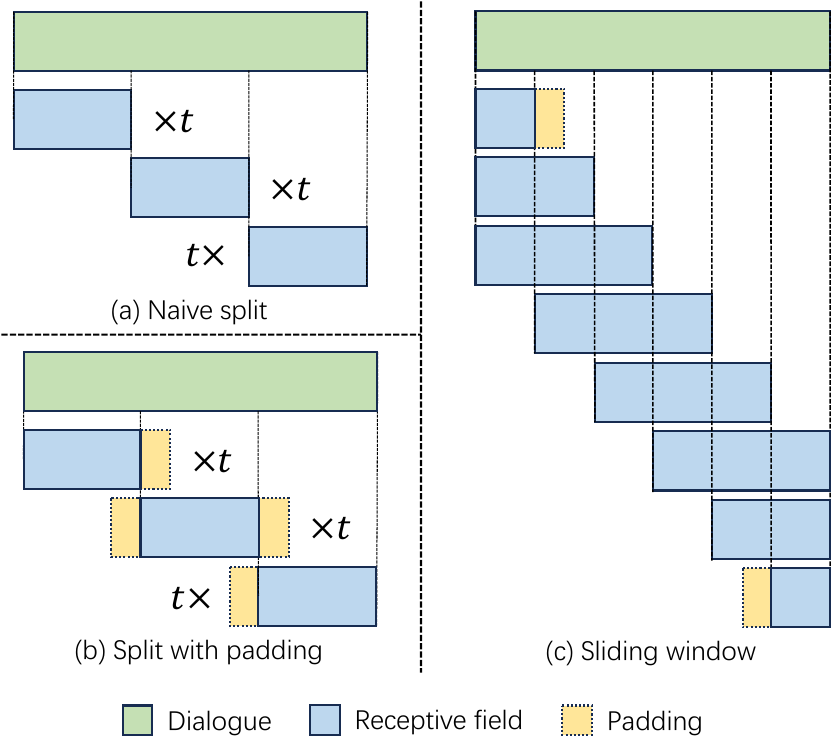}
    \caption{Strategies to split a dialogue: Figure \ref{fig:split}a is the naive splitting, where each receptive field is not overlapped with each other. Figure \ref{fig:split}b pads some samples at the beginning and the end of each receptive field. The $\times t$ and $t\times$ mean that repeat this receptive field for $t$ times. Figure \ref{fig:split}c demonstrated the sliding window strategy when $t=3$, which provides different receptive fields for each sample, mining dialogue features of different perspective views.}
    \label{fig:split}
\end{figure}
In practice, a dialogue could be too long to fit in the context window of the LLMs. Therefore, we had better split the dialogue into smaller windows. 
Another reason to split the contents is that the chances of passing the integrity check are negatively related to the number of samples passed to the LLMs. 
In addition, LLMs have output limits. If we feed too many samples in one prompt, they cannot assign enough thoughts to each sample and reach the limit before finishing adjusting all samples in the prompts. 
A simple strategy is to split the dialogue naively and repeat each window $t$ times as in Figure \ref{fig:split}a. 
However, a window may require some information from adjacent windows. 
We can pad dummy samples, like in Figure \ref{fig:split}b, so that each window could overlap with each other. 
Due to the randomness of the prediction results of LLM, we need to fuse the multiple predictions. 
To build a dimensionally uniform fusion algorithm, each sample will be covered in exactly $t$ windows and the paddings will be discarded. 
Nonetheless, for a sample, its predictions are all based on the same receptive field of a certain length. 
We hope that the predictions can include as many samples into consideration as possible. 
Therefore, we adopted a sliding window method to iterate the dialogue. 
During sliding, each prediction for the same sample is based on a different receptive field of the dialogue, which achieves feature extraction for the content of dialogues of different lengths. 
The Figure \ref{fig:split}c demonstrates an example when $t=3$. 
We can see that the window uses $\frac{1}{t}$ of its maximum length as the step size. 
It begins with containing only the amount of samples of one step, moves forward one step each time, and ends until the entire window is moved out of the dialogue range. 
The first and last windows have the least amount of data, we performed padding using dummy samples to prevent their context is not sufficient for LLMs to understand. 

No matter how we split the dialogues, we want the LLMs to have a comprehensive view of the entire dialogue. 
To achieve this goal, we ask the LLMs to summarize the dialogue and list some necessary ethical knowledge or common sense that might help identify the emotions before any adjustment request. 
This summary will be attached at every window for the LLMs to keep a global sight. 
\subsection{Receptive-Field-Aware Weighting} \label{sec:merge}
After passing the prompts to the LLMs, we can acquire a new prediction matrix $x\in\mathcal{R}^{n\times (t+1)}$ for each sample, where $n$ is the number of emotion classes and the $t+1$ represents $t$ predictions from the LLMs and one from the vanilla models. 
When combining these prediction results to make the final decision, the most naive strategy is simply adding up the $t+1$ predicted probabilities (or $t$ when ignoring the one produced by the vanilla models). 
However, this would ignore the internal patterns between each receptive field. We would like to learn the weights for them instead of assigning some arbitrary constants. 
Another proposal is to learn an $n\times (t+1)$ weight matrix and perform an element-wise product before summing up for each class. 
It ignored the length of the receptive fields for different predictions could be different. 
The lengths of receptive fields are different for the samples at the beginning or the end of the dialogue. 
In contrast, the samples in the middle would have the same length for their receptive fields. 
Here, we propose a receptive-field-aware attention-driven merge algorithm. 
$l\in\mathcal{R}^t$ is the proportion of the lengths of $t$ receptive fields with respect to the dialogue. 
First, the coarse importance of each prediction is calculated from the lengths of their receptive fields, denoted as $l'$, 
\begin{equation}
    l'=MLP_{t\times (t+1)}(l)\in\mathcal{R}^{(t+1)\times 1}
\end{equation}
Instead of training the entire projection matrices as parameters, we generate them from the lengths of the receptive fields. 
Specifically, 
\begin{equation}
    \begin{gathered}
            W_Q = MLP_{1\times (t+1)}(l')\in\mathcal{R}^{(t+1)\times(t+1)}\\
            b_Q = MLP_{1\times n}(l')^T\in\mathcal{R}^{n\times(t+1)}\\
            Q = xW^T_Q+b_Q\in\mathcal{R}^{n\times(t+1)}
    \end{gathered} 
\end{equation}
The rows of $x$ are the predictions for each emotion. 
The $Q_{i, j}$ is the linear combination of the $i$-th row of $x$, which is the importance of the $j$-th prediction in emotion $i$ considering all models' predictions in this emotion. 
The $i$-th row of $Q$ describes the importance of the predictions of emotion $i$, which can represent the importance of emotion $i$. 

In the calculation of $K$, the $x^T$ is used as the input instead of $x$ as in $Q$. 
The $x$ focuses on the predictions for each emotion, while the $x^T$ emphasizes the predictions of each receptive field. 
The $i$-th row of $K$ is $x^T_{i,:}W_K + b_{K_i}$. We can see it as the prediction pattern of the $i$-th receptive field summarized from predictions for all emotions of it. 
\begin{equation}
    \begin{gathered}
            W_K = MLP_{1\times n}(l')\in\mathcal{R}^{(t+1)\times n}\\
            b_K = MLP_{1\times (t+1)}(l')^T\in\mathcal{R}^{(t+1)\times(t+1)}\\
            K = x^TW^T_K+b_K\in\mathcal{R}^{(t+1)\times(t+1)}
    \end{gathered} 
\end{equation}

$V$ can be viewed as the first adjustment of each prediction considering the predictions of other receptive fields. 

\begin{equation}
    \begin{gathered}
            W_V = MLP_{1\times (t+1)}(l')\in\mathcal{R}^{(t+1)\times(t+1)}\\
            b_V = MLP_{1\times n}(l')^T\in\mathcal{R}^{n\times(t+1)}\\
            V = xW^T_V+b_V\in\mathcal{R}^{n\times(t+1)}
    \end{gathered} 
\end{equation}

The $W_{i, j}$ is the relative confidence between the $i$-th emotion and receptive field, which considered all predictions for $i$-th $Q_{i,:}$ emotion and all predictions of $j$-th receptive field $K_{j,:}$. 
\begin{equation}
        W = Softmax(\frac{QK^T}{\sqrt{t+1}})\in\mathcal{R}^{n\times (t+1)}
\end{equation}
The predictions are weighted by performing element-wise product of the weights and $V$. 
\begin{equation}
    x'=W\odot V\\
    y_i = \sum x'_{i,:}
\end{equation}


\section{Experiments}
\label{sec:exp}
We evaluated the performance of \sysname{} on the IEMOCAP \cite{busso2008iemocap} dataset. 
Since our method requires the classification label, dimension scores, and transcription, to the best of our knowledge, IEMOCAP is the only dataset that meets these requirements. 
We first discuss the effectiveness of using dimension scores to supplement information for classification. Then, introduce how we train our model. Finally, the detailed results of micro benchmarks and ablation studies are presented. 

\subsection{Dimension Scores and Emotional Classes}
\label{sec:dim_emo}


\begin{figure}[h]
    \centering
    \includegraphics[width=\linewidth]{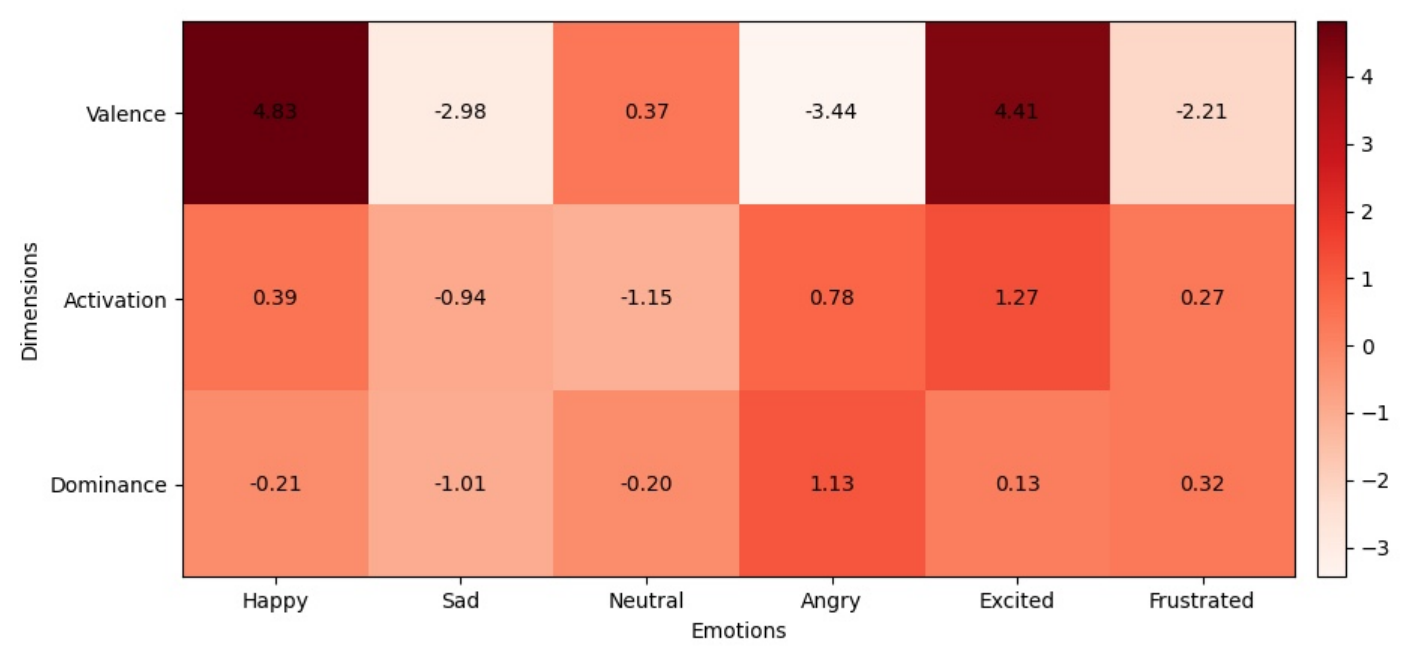}
    \caption{LDA coefficients}
    \label{fig:heat_map_dim_emo}
\end{figure}


\begin{figure}[ht]
    \centering
    
    \subfloat[Precision][\label{fig:precision_cm}]{%
        \includegraphics[width=0.48\linewidth]{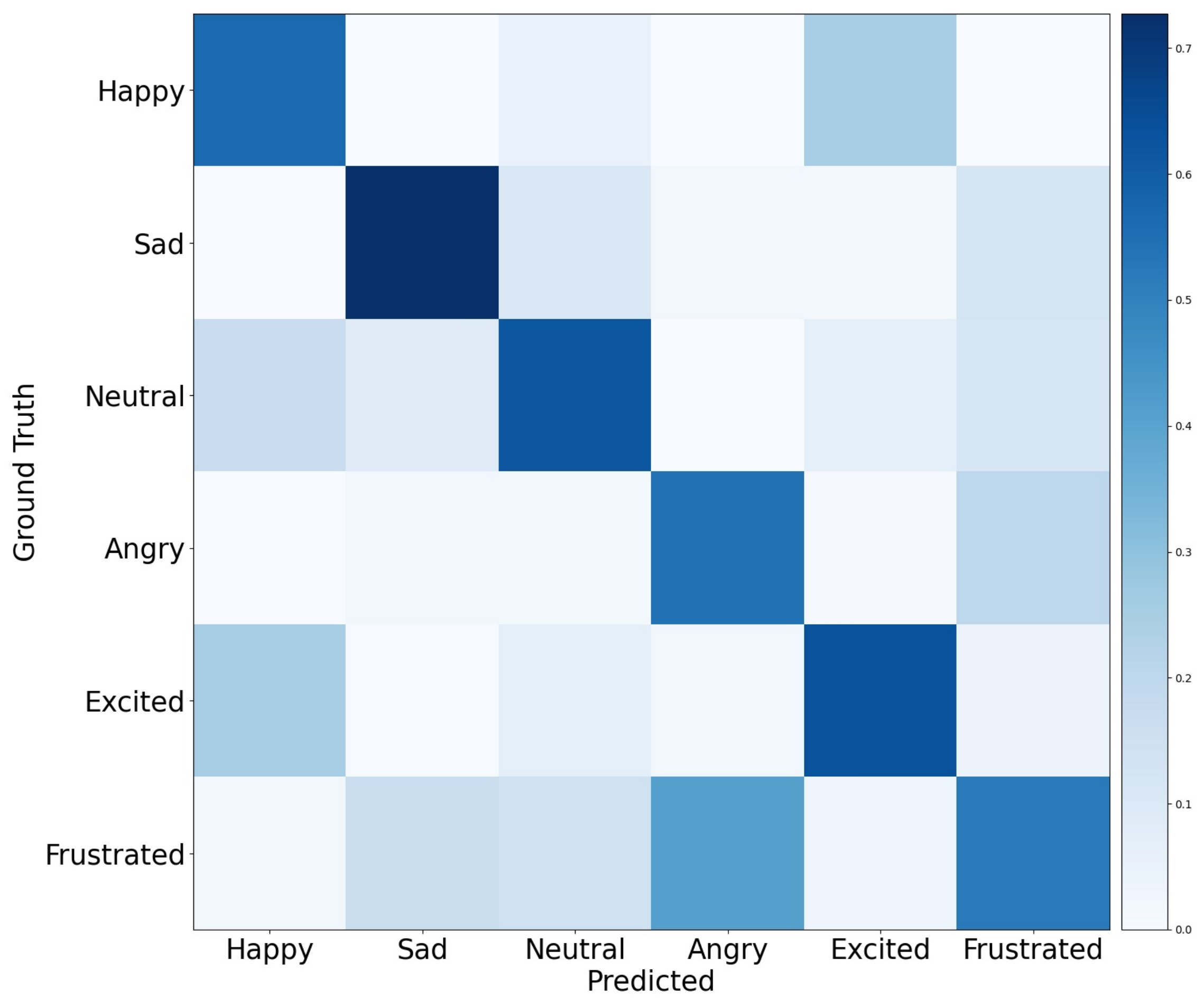}%
    }
    \hfill
    \subfloat[Recall][\label{fig:recall_cm}]{%
        \includegraphics[width=0.48\linewidth]{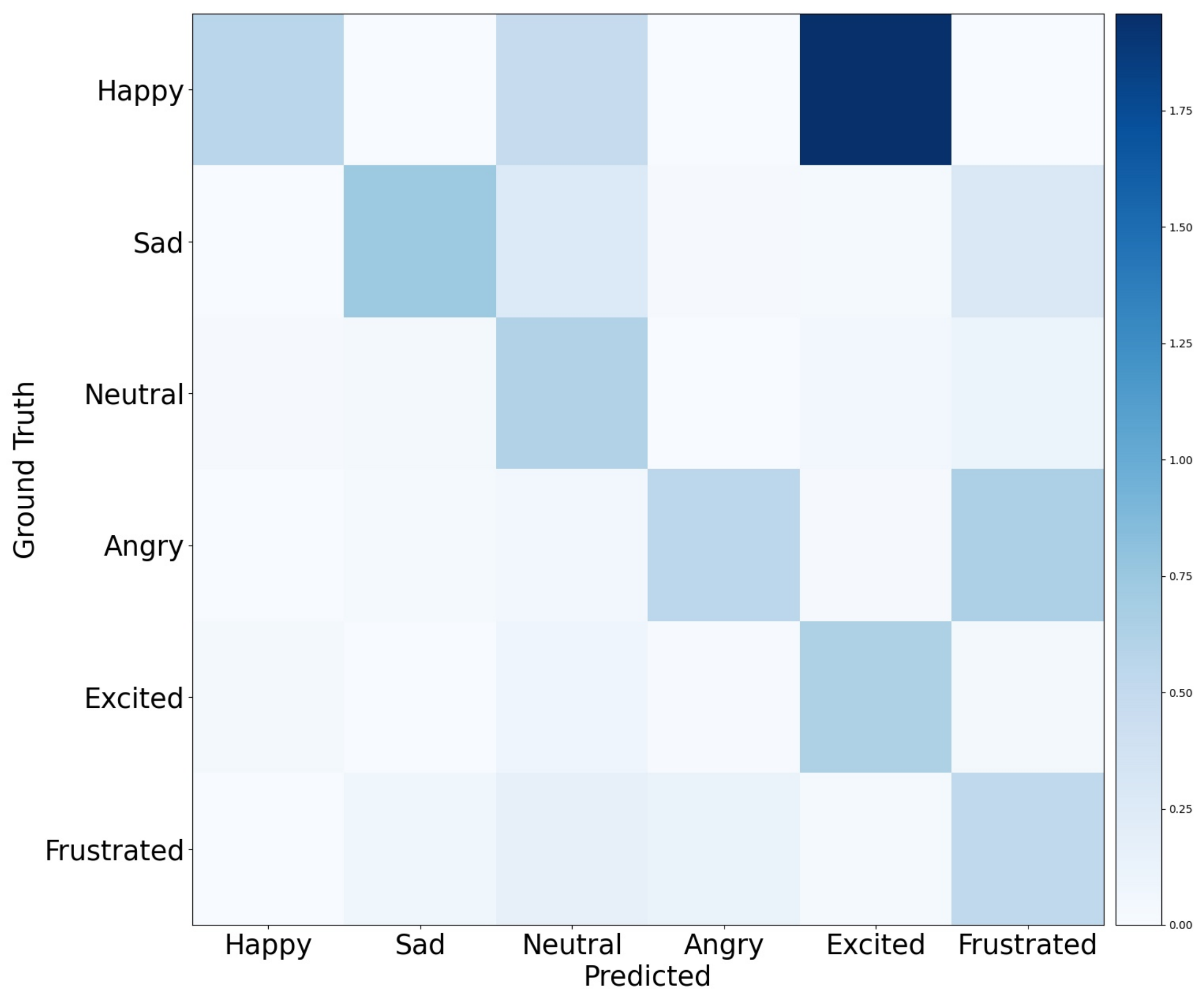}%
    }
    \caption{Confusion matrices of LDA results: panel~\ref{fig:precision_cm} is normalized by the sum of predictions, focusing on precision, and panel~\ref{fig:recall_cm} is normalized by the sum of ground truth, focusing on recall.}
    \label{fig:cm_dim_emo}
\end{figure}

\begin{table*}[ht]
\centering
\small
\caption{The results on IEMOCAP (6-way): all models are trained with multimedia modalities (acoustic, video, and transcription). The \textbf{bold} results are the highest scores in that column, while the \underline{underlined} ones are the second best.}
\label{tab:6class}
\resizebox{\textwidth}{!}{%
    \begin{tabular}{r|cc|cccccc}
        \hline
        \textbf{Models} & \textbf{Acc. (\%)} & \textbf{w-F1 (\%)} & \textbf{Happy} & \textbf{Sad}   & \textbf{Neutral} & \textbf{Angry} & \textbf{Excited} & \textbf{Frustrated} \\ \hline
        DialogueCRN & 65.31 & 65.34 & 51.49 & 74.54 & 62.38 & 67.25 & 73.96 & 59.97 \\
        COGMEN & 67.60 & 68.20 & 55.76 & 80.17 & 63.21 & 61.69 & 74.91 & 63.90 \\
        DialogueLLM & 70.62 & 69.93 & - & - & - & - & - & - \\
        CORECT & 69.93 & 70.02 & 59.30 & 80.53 & 66.94 & \underline{69.59} & 72.69 & 68.50      \\
        SDT & 73.95 & 74.08 & \textbf{72.71} & 79.51 & 76.33 & \textbf{71.88} & 76.79 & 67.14 \\
        \hline
        CORECT (Multi-task) & 70.18 & 70.28 & 57.64 & 86.11 & 68.00 & 64.42 & 79.11 & 64.43 \\
        CORECT + GPT-4  & 71.16   & 70.90    & \underline{70.10} & 86.22 & 64.76   & 65.02 & 76.28   & 68.12     \\ 
        \hline
        SDT (Multi-task) & 73.94 & 74.14 & 60.56 & \textbf{86.61} & 71.65 & 65.45 & \textbf{85.56} & \textbf{71.07} \\
        SDT + GPT-4 & \underline{74.43} & \underline{74.63} & 61.11 & \underline{86.55} & \underline{76.49} & 65.43 & \underline{85.02} & 68.37 \\
        SDT + Llama-405B & \textbf{74.73} & \textbf{74.79} & 63.41 & 81.20 & \textbf{78.78} & 64.92 & 82.39 & \underline{70.51} \\
        \hline
    \end{tabular}%
}
\end{table*}

We inspected the relationship between the dimension scores and the emotion classifications with the samples in the training set of IEMOCAP. 
A linear discriminant analysis is conducted, which achieved an overall precision of 60\% and recall of 54\%. 
Figure \ref{fig:heat_map_dim_emo} displays the LDA coefficients. 
They suggest that valence is a strong predictor, especially for happy and excited with high positive weights and sad and angry with strong negative weights. 
Arousal and dominance also play significant roles, with arousal being crucial for distinguishing angry and excited, and dominance for angry and frustrated. 

The confusion matrix is shown in Figure \ref{fig:cm_dim_emo}. 
Sad has the highest precision while excited and neutral have the highest recalls. 
Happy and excited are the most confusing classes. The high recall of excited is the result of the tendency to classify both happy and excited as excited. 
Angry and frustrated are another pair of confusing emotions. 


\subsection{Training Settings} \label{sec:training}

In our experiments, we adopted the CORECT \cite{nguyen2023conversation} and SDT \cite{ma2023transformer} as the vanilla models. To make these models support multi-tasking, we added 8 lines of code to the network of CORECT, 21 lines to the network of SDT, and around 50 lines to their training procedure to save results for the LLM adjustment. In each modification, we preserved the training method of the classification task. Specifically, in CORECT, a simple linear layer is appended to predict the dimension scores accepting the same feature as the input of the classifier, while, in SDT, self-distillation is also applied to the dimension scores by replacing the KL-diverse with MSE. 
The \textit{gpt-4-0125-preview} and \textit{Llama-3.1-405B} were the LLM deployed in our framework. 
Our settings completely followed previous studies.
The fifth session in IEMOCAP is used as the test set and the validation and training sets are split from the remaining sessions. We train the CORECT model on the training set and select the best model based on the validation set performance. The LLM then makes predictions on both the validation and test sets. The attention-based merge algorithm is trained on the validation set predictions, and the highest-scoring model is used for final test set predictions.

Our framework only introduced a prediction head and merge algorithm of the total parameter number of $(t+3)\times(t+1)+3n+3\times$feature\_size, where feature size is the output size of the backbone. 
The backbone of CORECT has around $4.78M$ parameters, including $300$ for the new prediction head for dimension scores. 
The CORECT and merge algorithm can be trained on a single NVIDIA RTX 3090 GPU, while the GPT is run on the cloud server of OpenAI. 

During the training, the loss function of the classification is cross-entropy loss, while the dimension scores use the concordance correlation coefficient (CCC) score. 
The final loss is a weighted sum of these two losses, where weights are arbitrary constants for different dataset settings. 

\subsection{Micro Benchmarks} \label{sec:mircro}

\begin{table}[]
    \centering
    \begin{tabular}{ccc}
        \hline
        \textbf{Models} & \textbf{Acc.} & \textbf{w-F1} \\ \hline
        CHFusion & 76.59 & 76.80 \\
        COGMEN & 82.29 & 82.15 \\
        CORECT          & 84.73             & 84.64             \\
        \hline
        CORECT (Multi-task) & \underline{85.47} & \underline{85.44} \\
        CORECT + GPT-4            & \textbf{86.53}             & \textbf{86.53}\\
        CORECT + Llama-405B & \textbf{86.53}             & \textbf{86.53}
        \\ \hline
    \end{tabular}
    \caption{The results on IEMOCAP (4-way): The training settings are the same as the 6-way results. }
    \label{tab:4class}
    \vspace{-0.4in}
\end{table}

Table \ref{tab:6class} exhibits the results on the 6-way IEMOCAP setting with multimedia modalities. 
Simple multi-tasking training can improve the performance of CORECT with 0.25\% in accuracy and 0.26\% in the F1-score. 
The accuracy and F1-sccore of the SDT are almost the same. 
This suggests that by applying multi-tasking loss signals, the model might be able to learn to understand the inputs more accurately and extract more useful features. 
It is no surprise that multi-tasking won't decrease performance since the dimension scores describe emotions from another perspective. 
However, we believe that it can increase the performance due to some emotions having remarkable patterns in the view of dimension scores. 
A straightforward proof is a significant improvement of excited (6.42\% for CORECT and 8.77\% for SDT) and sad (5.58\% for CORECT and 7.10\% for SDT), which is consistent with the results of the LDA analysis above. 
However, the side effect is also obvious. The improvements in excitement are also accompanied by a decline in the performance of happy, which also demonstrates the tendency of classifying happy as excited as the LDA analysis. 
Finally, both angry and frustrated are worse for CORECT (decreased by 5.17\% and 4.07\%), while the SDT became worse for angry (decreased by 6.43\%) and performed better for frustrated (improved by 3.93\%). 
These emotions are too similar in their dimension scores, making it difficult for the model to learn the distinguishing features. The SDT would tend to classify these samples as frustrated. 

After we applied the entire \sysname{} framework on CORECT with GPT-4, the accuracy further increased 0.98\% from the multi-task model and 1.23\% from the CORECT, while the F1-scores increased 0.62\% and 0.88\% respectively. 
When we combined the SDT with GPT-4, the accuracy and F1-scores were improved by 0.48\% and 0.55\% from the SDT, while integrating it with Llama can improve by 0.78\% and 0.71\% respectively. 
The reason our framework's effect on the SDT is not as significant as the COREST is that the CORECT can estimate the dimension scores more accurately. When performing self-distillation to the dimension score, we simply copied the hyper-parameters of the classification due to a lack of resources to select a set of more reasonable values. 
Notably, with the help of our framework, the models became better at recognizing happiness.
The price is the decrease in the performance of the excited
However, the accuracy of angry and frustrated is still not as good as the vanilla models. 

\begin{table}[]\small
    \centering
    \begin{tabular}{ccc}
    \hline
    \textbf{Sub-Modules} & \textbf{Acc.} & \textbf{w-F1} \\
    \hline
    -w/o dimension scores &  85.68   &  85.67   \\
    -w/o summary of dialogue & \underline{85.79} & \underline{85.79} \\
    -w/o COT in examples & 85.58 & 85.57 \\
    -w/o examples & 85.68 & 85.67 \\
    Split dialogue with padding & 85.68 & 85.63 \\
    \hline
    \sysname{}   & \textbf{86.53}    & \textbf{86.53}   \\
    \hline
    \end{tabular}
    \caption{Results when removing different parts of the prompt on the 4-way setting: The first row still uses the multi-task model as the vanilla model. 
    The third and fourth rows pass all samples in one prompt. 
    }
    \label{tab:ablation_prompt}
    \vspace{-0.4in}
\end{table}

\begin{table}[]\small
    \centering
    \begin{tabular}{c|cc|cc}
    \hline
    \multirow{2}{*}{\textbf{Merge algo.}} & \multicolumn{2}{c|}{\textbf{6-way}} & \multicolumn{2}{c}{\textbf{4-way}} \\ \cline{2-5} 
     & \textbf{Acc.}         & \textbf{w-F1}         & \textbf{Acc.}        & \textbf{w-F1}        \\ \hline
    Naive add-up  & 70.30 & 70.46 & 85.26 & 85.25 \\
    Naive weights & 69.62 & 69.02  &       86.00       &  85.99         \\ 
    Attention w/o RFA & 69.75 & 69.94 & 85.05 & 85.03 \\
    Attention w/ RFA & \textbf{71.16} & \underline{70.84} & \underline{86.32} & \underline{86.32} \\ \hline
    \sysname{}   &   \textbf{71.16}     &  \textbf{70.90}       &    \textbf{86.53}     &    \textbf{86.53}  \\ \hline
    \end{tabular}
    \caption{Results when using different merge algorithms using the same prompts as \sysname{}: The first row adds all $t$ predictions. The second row trains an $n\times (t+1)$ weight matrix and product with the predictions in element-wise. The last two rows are traditional attention \cite{vaswani2017attention} using $x$ as $Q$ and $x^T$ as $K$, $V$.  }
    \label{tab:ablation_merge}
    \vspace{-0.15in}
\end{table}

Table \ref{tab:4class} presents the results on the 4-way setting of IEMOCAP. Since the SDT did not perform the 4-way experiment, we only applied the CORECT model in this setting. 
We can observe similar results in the 6-way setting. 
The multi-tasking can improve the performance by 0.74\% in accuracy and 0.8\% in the F1-score. 
Introducing the multi-tasking training has a more significant improvement in the 4-way setting than the 6-way setting. 
The reason is that the excited is similar to the happy and the frustrated is similar to the angry from the perspective view of dimension scores. 
When eliminating the excited and frustrated from the dataset, the dimension score can provide a more distinguishable guide for classifications. 
The overall pipeline improved by 1.06\% in accuracy and 1.09\% in F1-score from the multi-task vanilla model, while 1.80\% and 1.89\% from the single-task vanilla model respectively. 
The improvements from the multi-task vanilla model are similar to the 6-way setting. 

\subsection{Ablation Studies} \label{sec:ablation}

In ablation studies, we measured the performance of removing different parts of our prompt and the alternative designs mentioned previously using CORECT as the vanilla model. 

Table \ref{tab:ablation_prompt} presents the ablation results for different parts of the prompt. 
Omitting any component of the prompt leads to a decrease in performance. 
Particularly, removing the chain-of-thought (COT) from the example answers shows the most significant negative impact. 
Interestingly, completely removing the examples performs better than removing only the COT. 
This discrepancy may stem from the selected examples not closely aligning with the samples, which confuses the LLMs when learning surface patterns from them. 
In contrast, the COT helps LLMs understand the reasoning process of a sample, thereby yielding more positive effects than unrelated examples.
Another notable observation is that when the COT is removed, the pass rate of the integrity check drastically increases. 
Even when all samples are consolidated into a single prompt, generating compliant responses becomes easier. 
We hypothesize this occurs because LLMs prioritize producing responses that fit the required format over reasoning for correct answers.
A related observation is that attempts to use \textit{GPT-3.5} with COT resulted in few responses passing the integrity check. 
Its ability seems constrained such that when focused on reasoning steps, significant errors occur in both results and format. 

Table \ref{tab:ablation_merge} shows the results using different merge algorithms. 
The first three settings that are unaware of the receptive fields show a notable decline in performance.
Some are even worse than the multi-task vanilla model. 
This underscores the importance of receptive-field awareness in the merge algorithm. 

Another phenomenon worth noting is that, when using SDT as the vanilla model, the accuracies of naively adding up the adjustment of the LLM are 73.51\% for GPT-4 and 72.95\% for Llama, which is 0.92\% and 1.78\% worse than using the receptive-field aware attention weighting while 0.43\% and 0.99\% worse than the multi-task model with LLM. The severely damaged performance is another evidence of the LLM being affected by the mistakes of the predicted dimension scores and the importance of our weighting method. 

\section{Discussion}

As demonstrated in the experiment section, our framework \sysname{} can effectively improve the performance of the vanilla multi-modal networks. 
As we demonstrated, any vanilla model, like CORECT \cite{nguyen2023conversation} or SDT \cite{ma2023transformer}, adding an extra prediction head to perform dimension scores and re-trained on the datasets that provided these labels can be deployed in our framework. 
Similarly, any power LLM can be leveraged as part of \sysname{}, including both more advanced frozen general LLMs, like GPT and Llama in our experiment, or the LLMs that are specifically fine-tuned for emotion recognition. 
We believe that the potential of \sysname{} goes far beyond what has been demonstrated in the experiments. 
We will keep exploring its potential in the future. 

The limitation of our framework is that it requires the vanilla model to be trained on both the classification and dimension scores. 
Collecting and labeling a high-quality dataset will consume a significantly large amount of resources and time. 
We have realized the importance of building a comprehensive and sufficiently large corpus, and we will continue to make contributions in this aspect. 

\section{Conclusion}

We present \sysname{}, a framework to improve the performance of vanilla models by prompting large language models with receptive-field-aware attention weighting. 
\sysname{} can be implemented with limited resources. Only a consumer-grade graphics card and a certain number of API invocations are needed. 
In different settings of IEMOCAP, we have proved that \sysname{} can significantly improve the performance of a vanilla model (CORECT in our work). 
We further analyzed the effect of multiple parts of our prompt and compared it with alternative algorithms to merge the predictions.

 \section*{Acknowledgment}

This work is supported in part by National Natural Science Foundation of China (No.61936015), Natural Science Foundation of Shanghai (No.24ZR1430600), XJTLU Research Development Funding (No.RDF-21-02-072) and Shanghai Key Laboratory of Trusted Data Circulation and Governance, and Web3.

\bibliographystyle{splncs04}
\bibliography{ref}

\end{document}